\documentclass[letterpaper, 10 pt, conference]{ieeeconf}
\IEEEoverridecommandlockouts

\usepackage[colorlinks]{hyperref}
\hypersetup{
    colorlinks,
    linkcolor=black,
    citecolor=black,
    filecolor=magenta,
    urlcolor=cyan,
}

\usepackage{resizegather}

\usepackage{cite}
\usepackage{amsmath,amssymb,amsfonts,mathrsfs}
\usepackage{algorithmic}
\usepackage{graphicx}
\usepackage{textcomp}
\usepackage{xcolor}
\usepackage{tcolorbox}

\usepackage{tablefootnote}
\usepackage{threeparttable}
\usepackage{caption, subcaption}

\usepackage{tikz}
\usepackage{graphicx}
\usepackage{tkz-euclide}

\usetikzlibrary{positioning}
\usetikzlibrary{shapes}
\usetikzlibrary{shapes.misc}
\usetikzlibrary{shapes.geometric}
\usetikzlibrary{plotmarks}
\usetikzlibrary{intersections}
\usetikzlibrary{calc}
\usetikzlibrary{fit}
\usetikzlibrary{patterns,tikzmark}
\usetikzlibrary{matrix,decorations.pathreplacing,calc}

\tikzset{cross/.style={cross out, draw, 
         minimum size=2*(#1-\pgflinewidth), 
         inner sep=0pt, outer sep=0pt}}

\newcommand{\vertiii}[1]{{\left\vert\kern-0.25ex\left\vert\kern-0.25ex\left\vert #1 
    \right\vert\kern-0.25ex\right\vert\kern-0.25ex\right\vert}}

\makeatletter
\renewcommand{\fps@figure}{htp}
\renewcommand{\fps@table}{htp}
\makeatother

\def\BibTeX{{\rm B\kern-.05em{\sc i\kern-.025em b}\kern-.08em
    T\kern-.1667em\lower.7ex\hbox{E}\kern-.125emX}}
    
\begin{document}

\title{Stoch BiRo: Design and Control of a low cost bipedal robot}

\author{GVS Mothish$^{1*}$, Karthik Rajgopal$^{2*}$, Ravi Kola$^{1}$, Manan Tayal$^{1}$, Shishir Kolathaya$^{1}$
\thanks{
This research was supported by the Pratiksha Young Investigator Fellowship and the SERB grant CRG/2021/008115.
}
\thanks{*The authors have contributed equally (arranged in alphabetical order)}
\thanks{$^{1}$Cyber-Physical System, Indian Institute of Science (IISc), Bengaluru.
{\tt\scriptsize \{mothishg, ravikola, manantayal, shishirk\}@iisc.ac.in}
}
\thanks{
$^{2}$ Birla Institute of Technology and Science(BITS), Pilani, Pilani campus.
{\tt\scriptsize \{f20190263P\}@alumni.bits-pilani.ac.in}}
.
}%

\maketitle
\begin{abstract}

 
 This paper introduces the Stoch BiRo, a cost-effective bipedal robot designed with a modular mechanical structure having point feet to navigate uneven and unfamiliar terrains. The robot employs proprioceptive actuation in abduction, hips, and knees, leveraging a Raspberry Pi4 for control. Overcoming computational limitations, a Learning-based Linear Policy controller manages balance and locomotion with only 3 degrees of freedom (DoF) per leg, distinct from the typical 5DoF in bipedal systems. Integrated within a modular control architecture, these controllers enable autonomous handling of unforeseen terrain disturbances without external sensors or prior environment knowledge. The robot's policies are trained and simulated using MuJoCo, transferring learned behaviors to the Stoch BiRo hardware for initial walking validations. This work highlights the Stoch BiRo's adaptability and cost-effectiveness in mechanical design, control strategies, and autonomous navigation, promising diverse applications in real-world robotics scenarios.
\end{abstract}


\section{Introduction}
\label{section: Introduction}
 \par
\begin{figure}[t]
    \centering
    \includegraphics[width=0.80\linewidth]{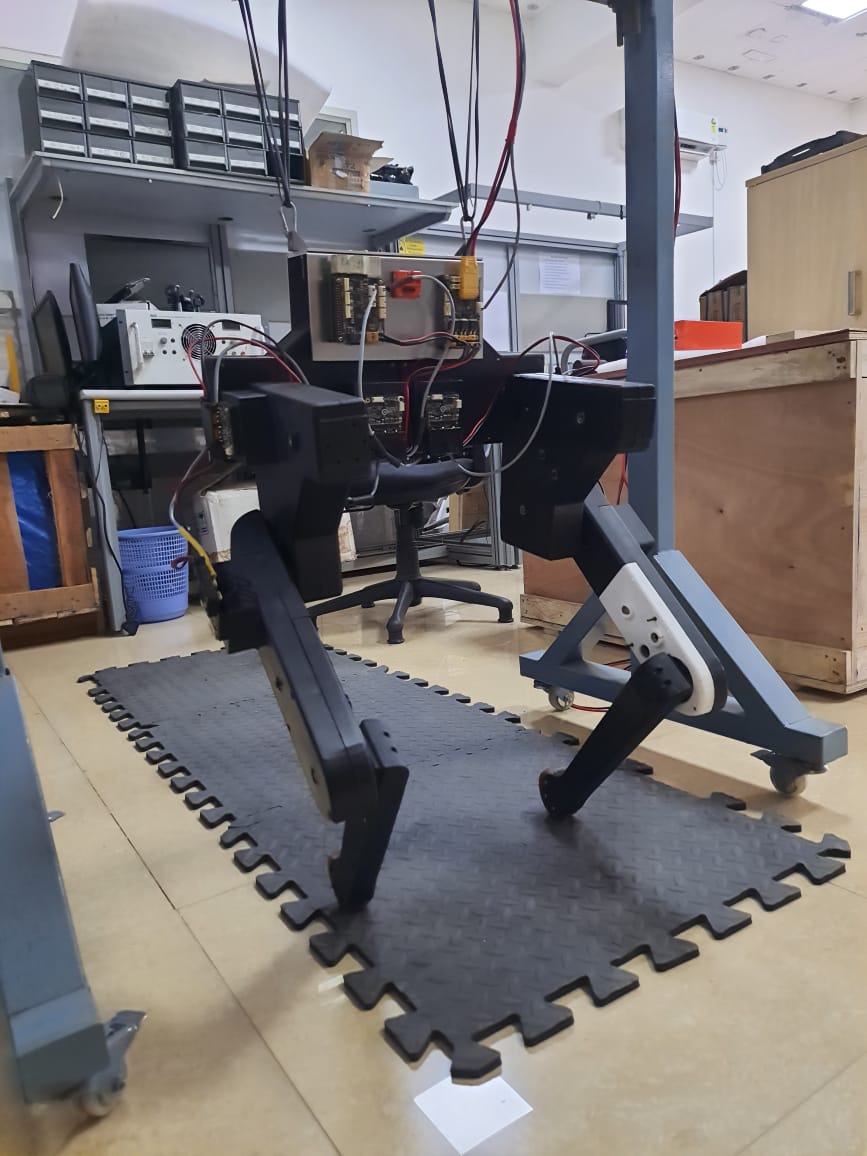}
\caption{Stoch BiRo: Hardware Testbed}
\label{fig:stoch-biro}
\end{figure}

 The realm of bipedal robotics has long been characterized by its potential for enabling machines to navigate environments akin to humans, yet its practical realization often faces challenges of high costs and limited adaptability to unstructured terrains. Addressing these limitations, this paper presents the Stoch BiRo, a pioneering low-cost bipedal robot shown in Fig. \ref{fig:stoch-biro}, specifically engineered to excel in traversing unfamiliar and uneven landscapes. With a deliberate focus on simplicity and affordability, the Stoch BiRo integrates a modular mechanical framework, incorporating point feet, to facilitate uncomplicated control strategies while ensuring adaptability in negotiating diverse and uncharted terrains.


It is often difficult to test advanced control and learning algorithms for legged robots without significant hardware development efforts or maintenance costs. Present-day hardware is often mechanically complex and costly, different robot systems are hard to compare to one another, and many systems are not commercially available. To support rapid and broad progress in academic research, we believe that hardware, firmware, and
middleware must become inexpensive and relatively easy to reproduce and implement. Open-source blueprints of low-cost legged robot platforms like Tik-Tok \cite{tiktok}, Bolt \cite{bolt} and ours allow researchers to test and develop their ideas on a variety of robotic platforms with different morphologies. In addition, the performance and capabilities of robots produced from open-source blueprints can be directly compared across laboratories.  the only other open-source torque-controlled legged robot platform. Most of its parts are waterjet cut and relatively easy to produce. Other open-source (hobby) projects mostly use position-controlled servo motors, limiting their usage to advanced control and learning algorithms. Complex machining is typical in quadruped robots exhibiting high-performance force control such as Agility Robotics' Cassie \cite{cassie}, Apptronik's Draco \cite{Apptronik}, . For an open-source legged robot to be successful, it is necessary to minimize the number of parts requiring precision machining, thereby favoring off-the-shelf components over sophisticated custom actuation solutions. To achieve this goal, we leverage recent advances in inexpensive plastic 3D printing and high-performance brushless DC motors (BLDC) which are now widely available as off-the-shelf components. Furthermore, we can take advantage of the improvements driven by the mobile device market, including affordable miniature sensors, low-power and high-performance microcontrollers, and advanced battery technologies. Lightweight, inexpensive yet robust robots are particularly relevant when testing advanced algorithms for dynamic locomotion. Indeed, ease of operation and collaborative open-source development can accelerate testing cycles.

\begin{figure*}[t]
\centering
\begin{subfigure}[b]{0.4\textwidth}
  \centering
  \includegraphics[width=\textwidth]{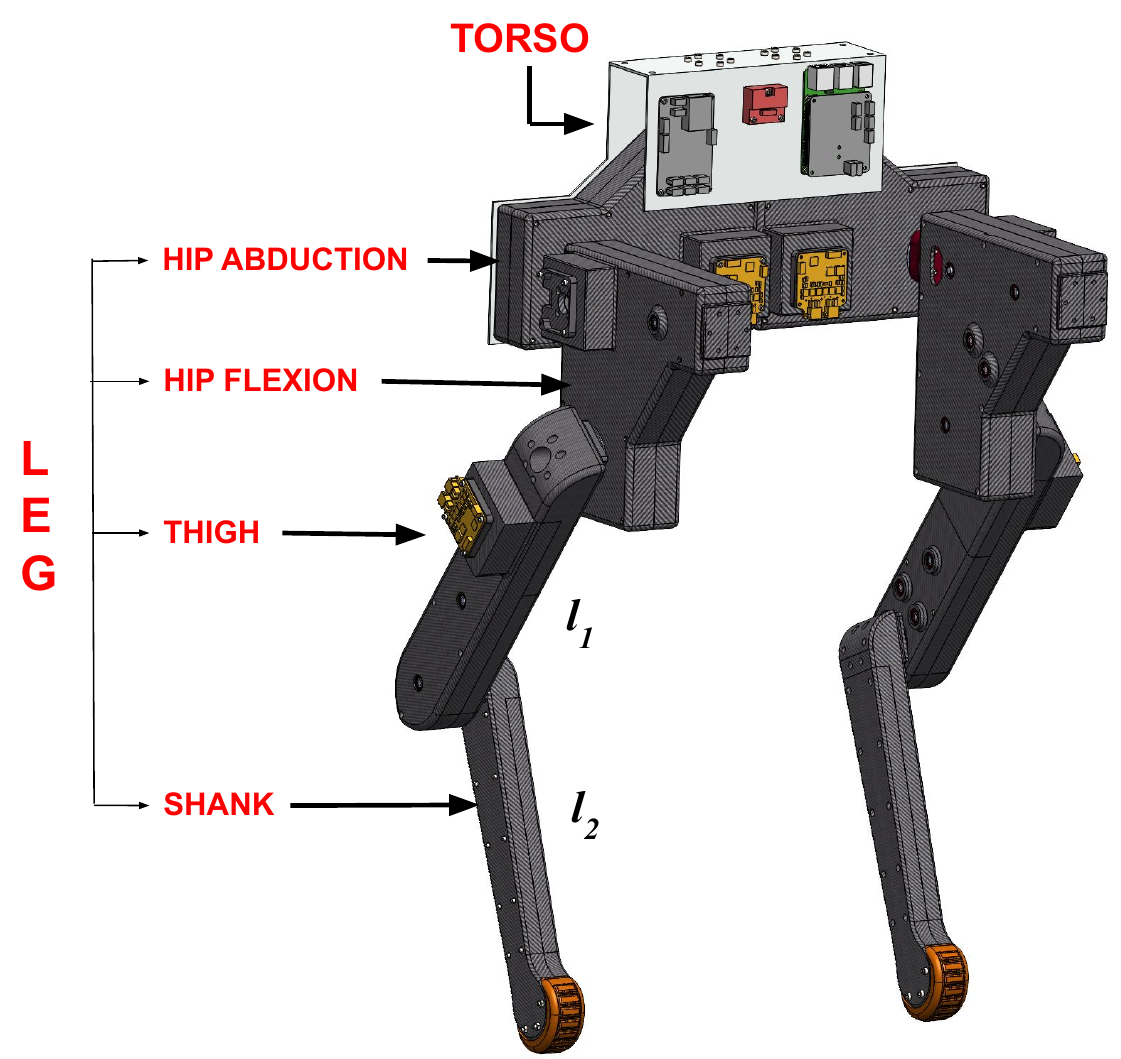}
  \caption{Master Assembly}
  \label{masterassem}
\end{subfigure}
\begin{subfigure}[b]{0.4\textwidth}
\centering
    \includegraphics[width=.8\textwidth]{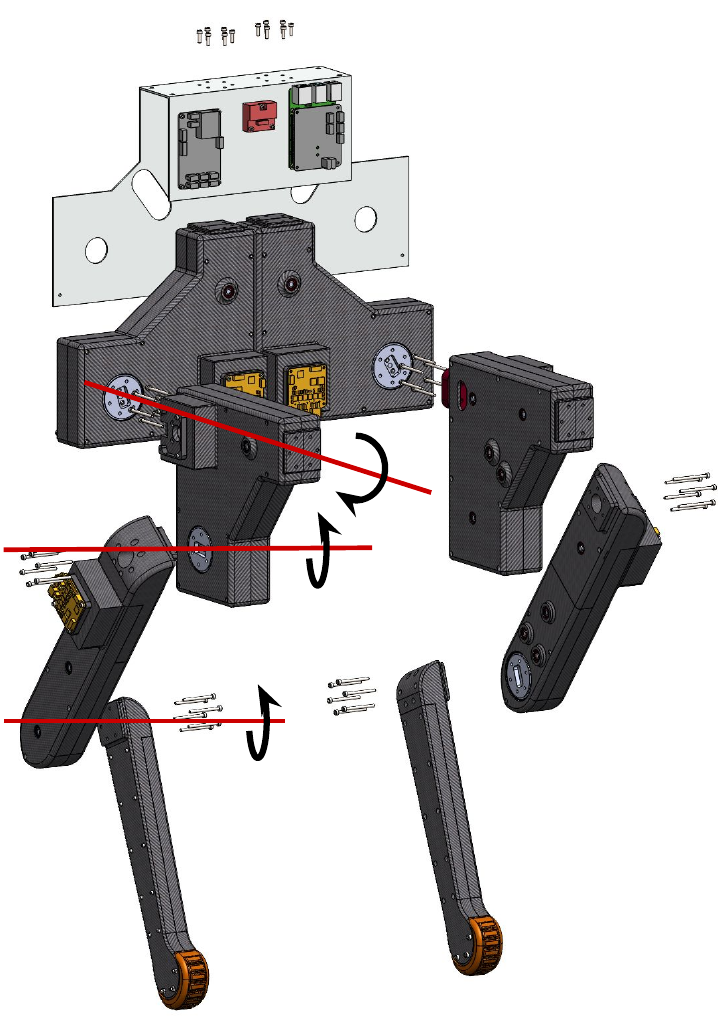}
    \caption{Exploded View}
    \label{exploded}
\end{subfigure}
\caption{CAD Assembly of Stoch BiRo}
\label{cadassem}
\end{figure*}
Model-based methods have proven effective in bipedal locomotion both by considering full models or by approximating simple models in designing the controllers. Approaches such as ZMP (zero moment point) \cite{park2001impedance,lim2004position,santacruz2013reactive} ensure that the center of pressure remains within the support polygon to achieve dynamic stability during walking. Research, such as \cite{raibert1985legged} presented the analysis and dynamics of walking on a single leg, Methods like \cite{apgar2018fast} combine the SLIP model with Raibert's approach coupled with MPC (Model Predictive Control) \cite{ferreau2008online}, LIP model \cite{kajita20013d}, Hybrid Zero Dynamics \cite{westervelt2003hybrid}, MPC with full order dynamics of model \cite{ding2022orientation}, Optimization-based \cite{tassa2012synthesis,kuindersma2016optimization,kuindersma2020recent} methods, achieved robust and stable walking by leveraging knowledge of the underlying model makes these kinds of methods depends on huge parameter tuning while extending to diverse and unknown terrains.

Reinforcement Learning (RL) based controllers \cite{sutton2018reinforcement} have gained popularity in rich locomotion behaviors \cite{heess2017emergence} \cite{benbrahim1997biped}. Some used Deep Reinforcement learning to train the robot \cite{xie2018feedback,siekmann2021sim,siekmann2021blind,li2021reinforcement,kumar2022adapting} thereby achieving novel and flexible walking trained in simulation and then by transferring to hardware. These methods generally require large-scale neural networks, with millions or even billions of parameters, and are computationally expensive to train and use. Additionally, their inherent nonlinearity makes it difficult to gain meaningful insights into their internal workings and leverage them for deeper understanding. This limits our ability to analyze and interpret their behavior, hindering further development and application.
Our approach is to combine model-based knowledge with RL framework \cite{castillo2022reinforcement,9682564}. Aiming for simplified controller development, and drawing inspiration from the successful implementation of linear policies in quadruped robots \cite{paigwar2021robust,rahme2021linear}, in bipedal robots with flat foot \cite{krishna2021learning,9682564} we present our work to extend this to low-cost point foot bipedal robot.

The main contributions of this paper are as follows: 

\begin{enumerate}
    \item a low-complexity, torque-controlled actuator module suitable for impedance and force control based on off-the-shelf components and 3D-printed parts.
    \item a computationally lightweight learning-based walking controller, demonstrating for the first time the execution of such motions on point-foot bipedal robots under moderate environmental uncertainty.
    \item an open-source platform comprising mechanical design files (STEP format), details of electronic components, and control software.
\end{enumerate}

\subsection{Organisation}

The rest of this paper is organized as follows. The robot design comprising the details of components used and actuation modules, along with the onboard computing platform, communication interfaces, and sensors are described in section \ref{section: Design}. A detailed description of the control architecture including the linear policy structure is explained in section \ref{section: Control Structure}. The simulation and experimental results are provided in section \ref{section: Results}. Finally, we present our conclusion in section \ref{section: Conclusions}.


\section{Robot Design}
\label{section: Design}
In this section, we first discuss the overall geometry of the mechanical structure, and computing systems, and then focus on its leg design and actuation principles.

\begin{figure*}[t]
\centering

\begin{subfigure}[b]{0.3\textwidth}
  \centering
  \includegraphics[width=\textwidth]{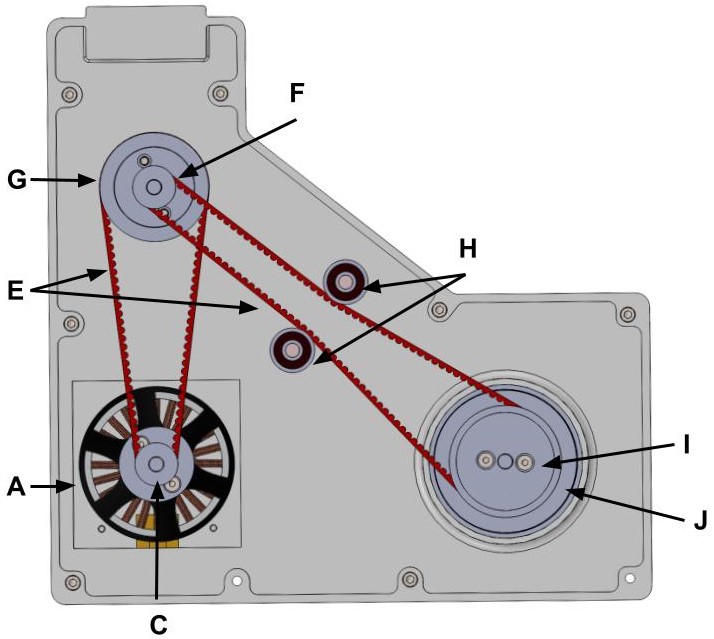}
  \caption{Abduction Module}
  \label{abduction}
\end{subfigure}
\begin{subfigure}[b]{0.3\textwidth}
\centering
    \includegraphics[width=\textwidth]{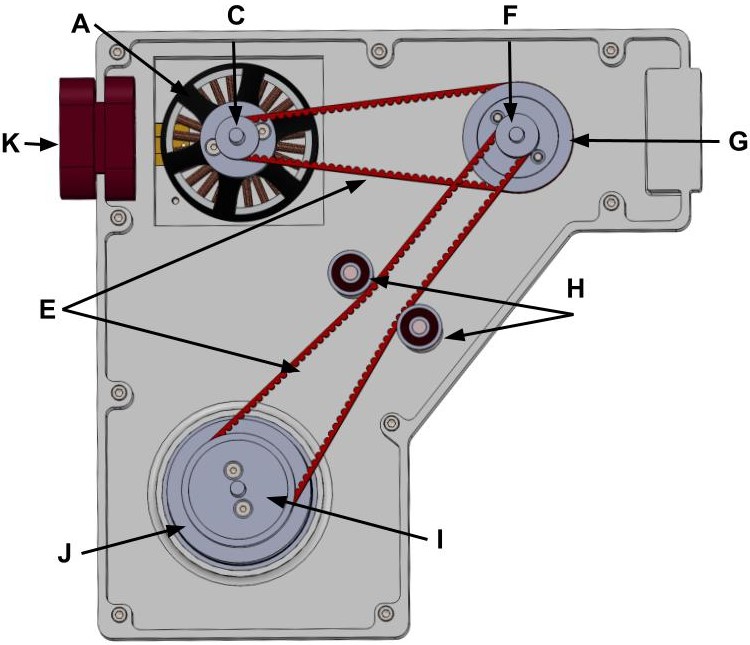}
\caption{Hip Module}
    \label{hip}
\end{subfigure}
\begin{subfigure}[b]{0.25\textwidth}
  \centering
  \includegraphics[width=\textwidth]{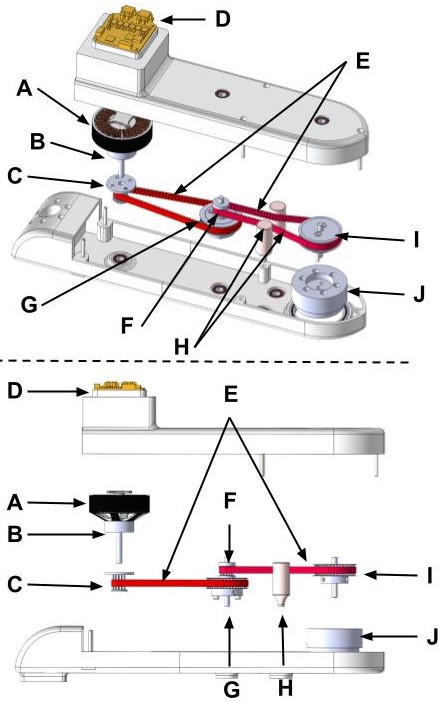}
  \caption{Thigh Module}
  \label{thigh}
\end{subfigure}
\caption{Actuator Modules (CAD). Components: A. T-Motor Antigravity MN5006 BLDC Motor, B. Motor coupling, C. Stage-I HTD Pulley (13 Teeth), D. MJBOTS Moteus r4.11 driver, E. HTD Timing Belts, F. Stage-II HTD Pulley (13 Teeth), G. Stage-I HTD Pulley (39 Teeth), H. Roller Tensioners, I. Stage-II HTD Pulley (39 Teeth), J. Joint coupling, K. Hip-to-hip attachment}
\label{modules}
\end{figure*}

\begin{table}[h]
    \centering
    \caption{Physical Robot Parameters}
    \begin{tabular}{c c c c}
    \hline
        Parameter & Symbol & Value & Units  \\
        \hline
         Mass & $m$ & 5.5 & kg \\
         Body Inertia & $I_{xx}$ & 0.0586 & kg . m$^2$ \\
                      & $I_{yy}$ & 0.0226 & kg . m$^2$ \\
                      & $I_{zz}$ & 0.0130 & kg . m$^2$ \\
         Body Length & $l_{body}$ & 0.46 & m \\
         Body Width  & $w_{body}$ & 0.28 & m \\
         Body Height & $h_{body}$ & 0.86 & m \\
         Leg link lengths & $l_1, l_2$ &0.27, 0.26 & m \\
    \hline
    \end{tabular}
    \label{robot_params}
\end{table}

\subsection{Geometry of Stoch BiRo}

The hardware design of Stoch BiRo builds on the actuation paradigm of Bolt biped from Open Dynamic Robot Initiative \cite{opendynamic}. The design philosophy behind Stoch BiRo is based on modularity, lightweight construction, ease of manufacturing, rapid repair, and reproduction. The robot design can be understood as an assembly of a central body module (torso) and two identical legs with three degrees of freedom in each leg. Fig. \ref{masterassem} illustrates the location of the modules in the biped and Fig. \ref{exploded} shows an exploded view of how these individual modules are connected. Stoch BiRo weighs around 5.5kg with a standing height of 0.75 m and an overall height of 0.84 m from the ground to the top of the torso. The robot's hip modules are laterally 0.33m apart and further design parameters are mentioned in \ref{robot_params}.

\subsection{Torso \& computing and communication}
The torso consists of two L-shaped aluminum plates bolted together, along with electronic components, cable routing, and a power distribution board. The power is supplied to the bipedal robot through an extension from a 6S LiPo Battery. The plates have been manufactured using laser cutting followed by sheet metal bending. It houses the central processing hub of the robot is a Raspberry Pi 4, functioning on a Linux-based operating system. This component utilizes its computational prowess for executing tasks related to locomotion control and state estimation. The processor communicates through a Pi3Hat daughter board to MJBOTS moteus r4.11 drivers on each actuator via a 5Mbps CAN interface with each leg in a single CAN bus. This setup executes control strategies, including joint PD control at 200 Hz. For inertial measurement, we use an Xsens MTi-610 IMU, which provides calibrated data on the 3-D orientation, angular velocities, and acceleration.

\begin{table}[h]
\centering
\caption{HTD Timing Belts Specifications}
\begin{tabular}{c c c c c c}
\hline
Sl. No. & Pitch & Width & Total Length & Belt Height & Tooth Height \\
\hline
1. & 3mm & 6mm & 279mm & 2.4mm & 1.17mm \\
2. & 3mm & 6mm & 399mm & 2.4mm & 1.17mm \\
\hline
\end{tabular}
\label{belttable_1}
\end{table}

\begin{table}[h]
\centering
\caption{HTD Pulley Specifications}
\begin{tabular}{c c c c c}
\hline
Sl. No. & Bore Diameter & Outer Diameter & No. of Teeth & Width \\
\hline
1. & 5mm & 16mm & 13 & 16mm \\
2. & 5mm & 44mm & 39 & 16mm \\
\hline
\end{tabular}
\label{pulleytable_1}
\end{table}

\subsection{Leg Design \& Actuation}

Each leg has three actuator modules: hip abduction/adduction, hip flexion/extension, and knee flexion/extension, with a shank attached at the end. All the modules consist of a brushless DC motor (T-Motor Antigravity MN5006 KV300; refer to Table \ref{motor_specs}) coupled to a dual-stage HTD timing belt actuation with a 3:1 reduction at each stage, resulting in an overall multiplication of 9 times the torque motor has to offer. The belts and pulleys specifications are provided in Tables \ref{belttable_1} and \ref{pulleytable_1}, respectively. Roller tensioners are provided to prevent any slack by consistently providing tension in the timing belts while in motion. The hip abduction and flexion modules are triangular to achieve the desired torque reduction (9:1) in minimal space, with the thigh module and shank being linear in structure. The leg modules' coverings, hip-to-hip attachments, tensioners, pulleys, and joint couplings have adopted Fused Deposition Modelling (FDM) 3D printing for their manufacturing. The motor couplings and the shafts (5mm diameter) for the pulleys have been machined using turning operation. The rest of the components (timing belts, bearings, motor, and motor drivers) have been procured off the shelf. The modules are configured such that the wiring from the Moteus drivers does not interfere with the motion of the leg modules as the wires would be entangled during its motion which could disengage the electrical connections as well as the robot to be tripped. Also, the inward orientation of the shank to other modules will enhance stability in its gait. The emphasis in the leg design was to keep the inertia of the moving segments minimal. A detailed view of the leg assembly is shown in Figure \ref{masterassem} and key specifications of the actuators are summarized in Table \ref{actuators_params}. The descriptions of the leg links are as follows:

\begin{enumerate}
    \item \textbf{Abduction Module}: The abduction module, as shown in Fig. \ref{abduction}, is the topmost module of a leg in the biped, which is responsible for the abduction/adduction movement in the frontal plane of the robot's hip, similar to that of humans. The module is attached to the torso plate and actuates the hip abduction-flexion coupling through the timing belts-pulleys system. This coupling is a part of the abduction joint, which connects the abduction and the flexion modules. 
    \item \textbf{Hip Flexion Module}: The hip flexion module (refer to Fig. \ref{hip}) is placed between the hip abduction and thigh modules. This module is mostly identical to that of the abduction module with few changes as it performs the flexion-extension motion in the sagittal plane. The hip-to-hip attachment in this module facilitates the transfer of actuation from the abduction motor. The output pulley is connected to a flexion-thigh coupling, forming the flexion joint, which connects the flexion and thigh modules.
    \item \textbf{Thigh Module}: The thigh module (illustrated in Fig. \ref{thigh}) is responsible for the knee flexion/extension of the biped, similar to the primary movement of human knees. The motor in this module controls the knee joint through the thigh-shank coupling. The output pulley is attached to this coupling and the motion is transmitted to the shank module. Unlike the previous two modules which had a triangular configuration, the belts of this module are arranged linearly.
    \item \textbf{Shank}: Shank is the lowest part of the biped leg, which runs from the knee to the foot. This module is similar to the structure of the thigh but without any of the actuating components (motors, belts, pulleys, etc.) and with a provision for foot attachment at its bottom end. A point contact foot made of polyurethane (PU) is attached to the end of the shank for better grip of the biped when in contact with a surface. 
\end{enumerate}

Further details on the mechanical design of Stoch BiRo are available on the open-sourced platform\footnote{\label{note: webpage} \url{https://tayalmanan28.github.io/Stoch-BiRo/}}.


\begin{table}[h]
    \centering
    \caption{Actuator Parameters}
    \begin{tabular}{c c c}
    \hline
        Parameter & Value & Units  \\
        \hline
         Gear Ratio & 9 & - \\
         Max Torque & 14 & N-m \\
         Max Joint Speed  & 610 & RPM \\
    \hline
    \end{tabular}
    \label{actuators_params}
\end{table}

\begin{table}[h]
    \centering
    \caption{Brushless DC Motor Specifications}
    \begin{tabular}{c c c}
    \hline
        Parameter & Value & Units  \\
        \hline
        Model & T-Motor Antigravity MN5006&\\
         Mass & 106 & g \\
         Motor Size & 55.6 (Outer Diameter) & mm \\
                    & 30 (Width)& mm\\
         Peak Current (180s) & 20 & A \\
         KV Rating & 300 & \\
         Maximum Power (180s)  & 500 & W \\
         Maximum Torque & 0.5 & N-m \\
    \hline
    \end{tabular}
    \label{motor_specs}
\end{table}

\section{Control Architecture}
\label{section: Control Structure}

\begin{figure*}
    \centering
     \includegraphics[width=1\linewidth]{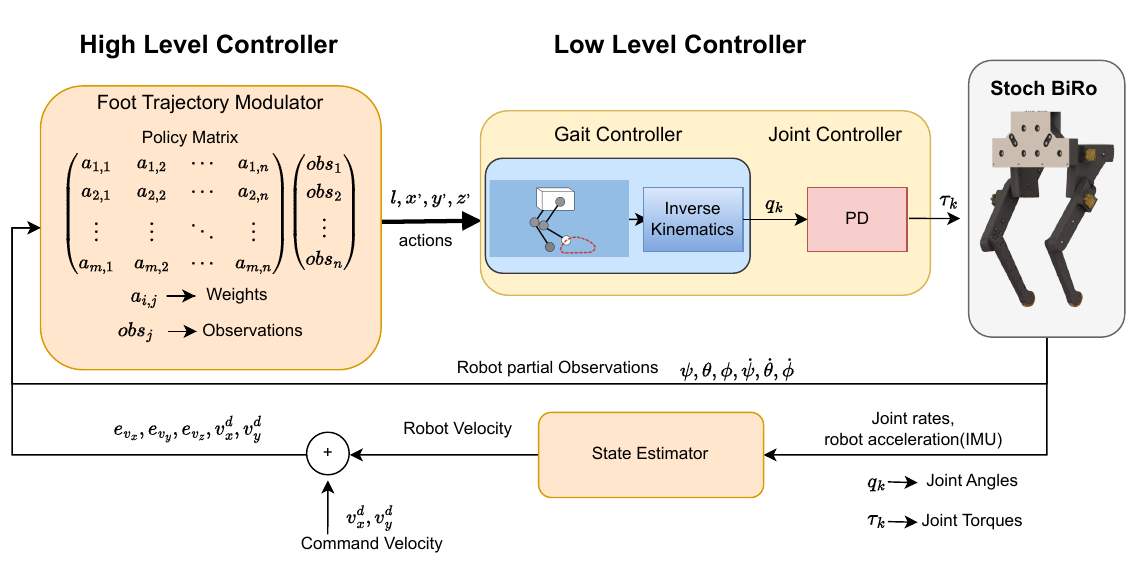}
\caption{Controller Architecture}
\label{fig:controller}
\end{figure*}

Our control architecture follows a hierarchical structure, similar to the one used in \cite{9682564}. We are explaining high and low-level controllers in the two subsequent sections as shown in the figure \ref{fig:controller}.

Here, the high-level controller analyzes the observed robot state and determines the parameters for the semi-elliptical foot trajectories and foot placement shifts along the x, y, and z axes. These parameters then act as inputs for the lower-level controller, which calculates the desired joint angles. Finally, the joint angles are tracked by a dedicated tracking system.

This hierarchical approach allows for a decoupling of the control process, simplifying the design and implementation. The high-level controller focuses on strategic decision-making based on the robot's overall state and terrain, while the lower-level controller handles the complex task of joint angle computation and tracking for achieving the desired foot trajectories.

\subsection{High Level Controller}
Approaching the locomotion challenge from a Reinforcement Learning (RL) perspective, we aim to develop a policy capable of parameterizing the trajectories of the robot's feet. These parameterized trajectories act as inputs for the gait controller, which, in turn, incorporates a regulatory mechanism based on the contact state of each leg.

In the specific context of a given state space \(S \subset \mathbb{R}^n\) with dimension \(n\) and an action space \(A \subset \mathbb{R}^m\) with dimension \(m\), we define our policy as \(\pi : S \xrightarrow{} A\). Mathematically, this is expressed as \(\pi(s):= Ms\), where \(M \in \mathbb{R}^{m \times n}\) is the Policy matrix containing parameters subject to learning. Emphasizing the intentional reduction of control complexity, we affirm that a linear policy proves sufficient for mastering such a transformation.

Further insights into the reasoning behind our selection of observation and action spaces are provided below.

\textbf{State Space}: In our earlier study\cite{paigwar2021robust,krishna2021learning,9682564}, we highlighted the efficacy of selecting a truncated observation space from the complete set of robot states. Consequently, the state space is characterized by an 11-dimensional state vector denoted as $s_t = \{\psi, \theta, \phi, \Dot{\psi}, \Dot{\theta}, \Dot{\phi}, e_{v_x}, e_{v_y}, e_{v_z}, v^d_x, v^d_y\}$. An invariant extended Kalman filter (InEKF)\cite{ekf} is derived for a system containing IMU with forward kinematic correction measurements for state estimation.

\textbf{Action Space}: The semi-elliptical trajectory of the foot is defined by the major axis, denoted as the Step Length ($l$), and translational shifts along the $X$, $Y$, and $Z$ directions represented by $\Acute{x}$, $\Acute{y}$, and $\Acute{z}$ (collectively denoted as $\Acute{O}$). In this context, the step length characterizes the walking motion, while the shifts play a crucial role in actively balancing the robot.

To maintain symmetry in the trajectories, we eliminate all asymmetric conventions between the legs beyond the policy and apply a mirrored transformation based on the leg. This approach allows us to learn and predict a single set of parameters regardless of the leg. Consequently, the action space is a 4-dimensional vector, expressed as $a_t=\{l,\Acute{x}, \Acute{y}, \Acute{z}\}$.

\subsection{Low Level Controller}
The low-level controller is subdivided into two parts, gait controller and motor level controller (joint controller). The gait controller generates the semi-elliptical foot trajectory using the parameters obtained from the high-level controller and disintegrates as per the control step frequency, thereby continuously generating joint angles for both swing and stance legs using inverse kinematics (each leg is modeled as a 3R manipulator in 3D space). A phase-variable $\tau\ \in [0, 1)$ which tracks the semi-elliptical trajectory gets reset once every walking step or upon a foot contact. Hence for an ideal walking cycle, the phase variable iterates from 0 to 1 twice.

The motor level controller tracks the motor positions from the gait controller by using a Proportional-Derivative (PD) controller with appropriate gains $K_p$ and $K_d$ as follows:
\begin{equation}
    \tau_{\text{fb}} = K_p (q_d - q) + K_d (\dot{q}_d - \dot{q}),
\end{equation}
where $q_d$ are the desired joint angles, $\dot{q}_d$ is the desired joint velocity, and $K_p$, $K_d$ are the proportional and derivative gains, respectively.

\section{Policy Training}
\label{section: Reinforcement Learning}
\begin{figure*}
    \centering
    \begin{subfigure}{0.48\linewidth}
        \includegraphics[width=\linewidth]{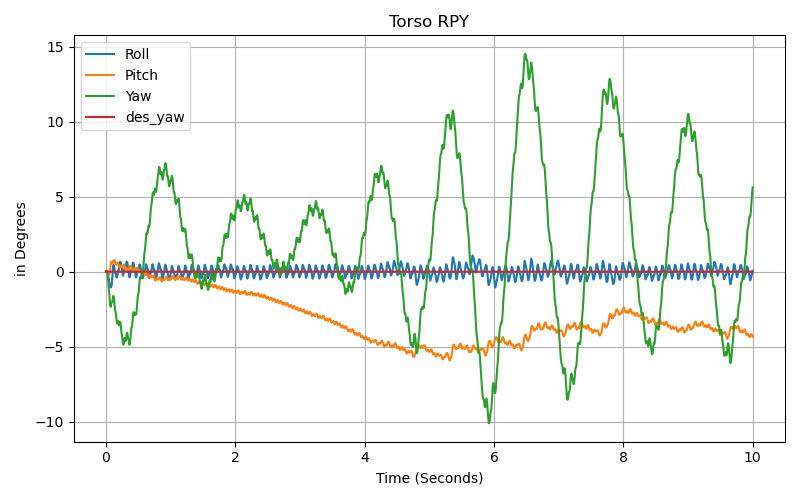}
        \caption{TorsoRPY for Flat Terrain}
        \label{fig:flattorsorpy}
    \end{subfigure}
    \hfill
    \begin{subfigure}{0.48\linewidth}
        \includegraphics[width=\linewidth]{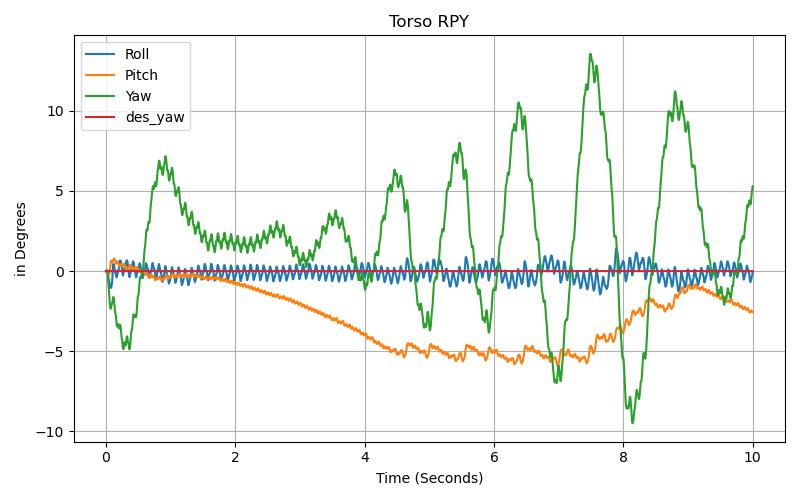}
        \caption{TorsoRPY for 3-degree slope}
        \label{fig:3degtorsorpy_slope}
    \end{subfigure}
    \hfill
    \begin{subfigure}{0.48\linewidth}
        \includegraphics[width=\linewidth]{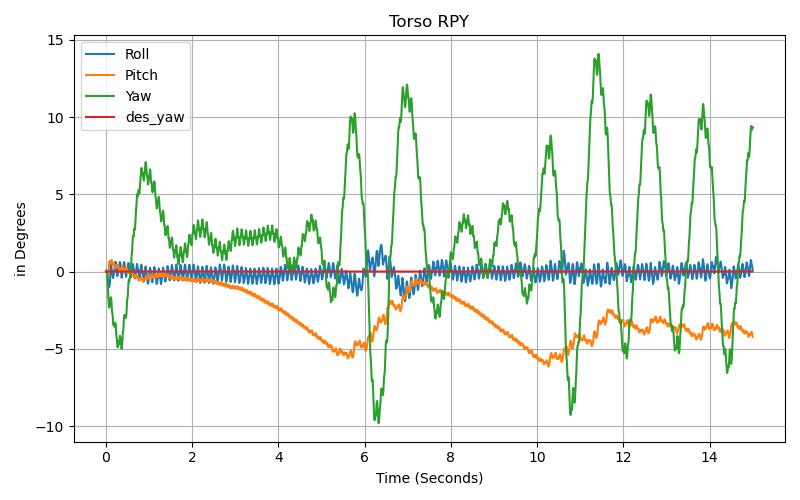}
        \caption{TorsoRPY for Sinusoidal Terrain}
        \label{fig:torsorpy_sinusoidal}
    \end{subfigure}
    \caption{Robot's RPY (Roll, Pitch and Yaw (in degrees)) on different terrains}
    \label{RPY_graphs}
\end{figure*}

This section details the training process for the learning based linear policy controller. We utilize Augmented Random Search (ARS) \cite{DBLP:journals/corr/abs-1803-07055} for optimization. Notably, instead of the conventional training setup with a search space of $\mathbb{R}^{m \times n}$, we exploit the heuristic structure of the policy matrix by using the prior knowledge of a model and explore a smaller sub-space within this parameter space.

\subsection{Reward Function}
Our Reward function consists of three parts, first part($w1,w2,w3$) ensures stability by penalizing robot roll, pitch and height from desired values. Second part($w4,w5$) ensures that robot will follow velocity commands. And final term rewards the robot for taking steps. The function is defined as,
\begin{equation}
\begin{split}
    r = G_{\omega1}(e_\psi) + G_{\omega2}(e_\theta) + G_{\omega3}(e_{p_z}) + \\ G_{\omega4}(e_{v_x}) + G_{\omega5}(e_{v_y}) +  W\Delta_x
\end{split}
\end{equation}
In the context of our study, $e_\psi$, $e_\theta$, $e_{p_z}$ denotes the deviation in the robot's roll, pitch, height from ground respectively, and $\delta x$ represents the displacement along the current heading direction during a specific time step, with a weighting factor $W$. $e_{v_x}$, $e_{v_y}$ denotes errors in velocities. The function $G: \mathbb{R} \xrightarrow{} [0,1]$ corresponds to a Gaussian kernel defined as $G_w(x) = \exp(-w \cdot x^2)$, where $w > 0$. Where as the height reward is ignored while training for slopes and sinusoidal terrains.

The primary objective in this scenario is to maximize the distance covered while ensuring the stability of the robot's torso.

\subsection{Training Setup}
We initiated the training with a hand-tuned policy to expedite the learning process. This approach is preferred due to its minimal hyperparameters, user-friendly implementation, and demonstrated efficacy in continuous-control tasks.

At the onset of each episode, a specific terrain configuration is randomly selected from a discrete set corresponding to that terrain. The target heading velocity is maintained at a small positive value to discourage the policy from learning to remain stationary. An episode concludes when any of the following conditions are met: (i) if the robot's torso surpasses predefined limits, (ii) if the robot's height falls below a specified threshold, or (iii) when the maximum episode length is reached.

The training was performed on Intel i7 processor with 16GB RAM for 4 to 5 hours. For training the policy, we employ ARS with the following hyperparameters: Learning rate ($\beta$): 0.01, Noise ($v$): 0.01, and Episode length: 10,000 simulation steps.

\section{Results \& Discussions}
\label{section: Results}
This section presents our simulation results trained on Mujoco \cite{todorov2012mujoco} physics engine with Gym environment and some initial hardware results. 
\subsection{Simulation Results}
\begin{figure}
    \centering
    \begin{subfigure}{1\linewidth}
        \includegraphics[width=\linewidth]{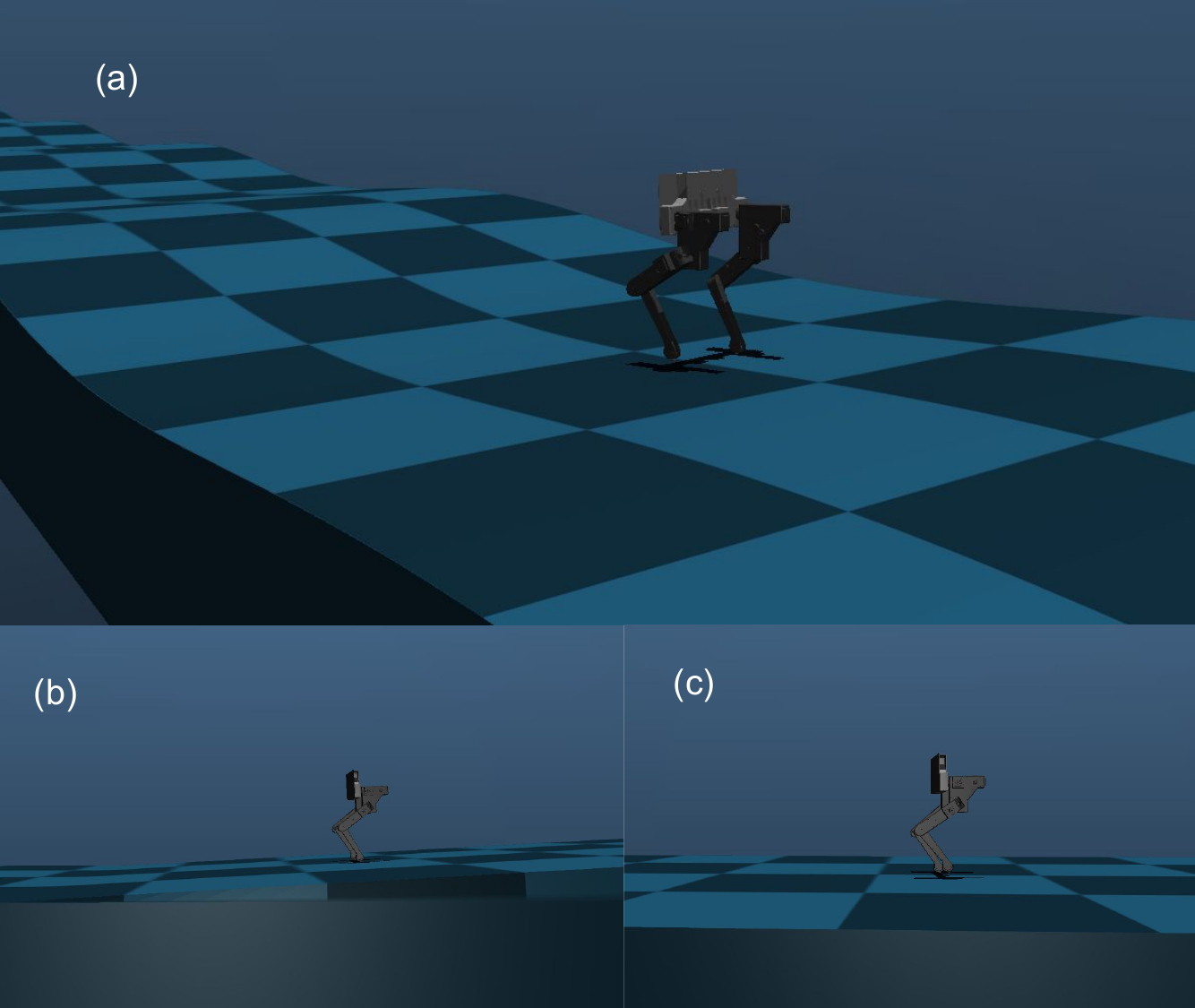}
    \end{subfigure}
    \caption{Simulation testbed on Flat Terrain, Slope terrain, Sinosoidal Terrain}
    \label{fig:simres}
\end{figure}

We conducted simulations under three different conditions as shown in figure \ref{fig:simres} to assess the robustness of the policy. The commanded velocity was set to 0.15 m/s in all figures. 

Figure~\ref{RPY_graphs} illustrates bipedal walking on various terrains and shows the Roll, Pitch, and Yaw (RPY) angles for the torso, measured in degrees for flat terrains as in figure \ref{fig:flattorsorpy}. For walking on slopes, figure~\ref{fig:3degtorsorpy_slope} showcases the results. The maximum yaw disturbances reached during walking on even steeper slopes was 10.5 degrees. In the 3-degree slope setting, the robot experienced a slip and destabilization, but the policy took some time to regain stability and resume walking, as indicated in the figure. The robot's height and bent legs constrained its ability to navigate steeper slopes. figure~\ref{fig:torsorpy_sinusoidal} presents data for the robot walking on a sinusoidal terrain. In all the graphs, the policy effectively minimized deviations in roll and pitch, likely due to the reward function. Yaw constraints were not rewarded for avoiding constrained and unnatural walking.

\subsection{Hardware Results}
The trained policy is transferred from simulation to Hardware with minimal parameter tuning. Our demonstrations include flat terrain walking in laboratory setup. Explore the video on the webpage$^{\ref{note: webpage}}$.


\section{Conclusions}
\label{section: Conclusions}
This paper introduces the Stoch Biro, a cost-effective point foot bipedal platform accompanied by computationally efficient linear policy based control framework. The presented model highlights proficient walking abilities with low computational demands and a lightweight hardware design. The demonstrated robustness, achieved without relying on external sensing, underscores its capability for successful locomotion in challenging situations.

Future hardware improvements are geared towards handling increased torque, expanding the robot's operational capabilities for traversing challenging terrains, and carrying battery pack weight. Simultaneously, efforts to mitigate yaw disturbances involve refining control algorithms and sensor fusion methods. Additionally, we plan to optimize velocity tracking for improved task planning, focusing on enhancing the robot's responsiveness and precision in executing complex maneuvers including, ensuring adaptability in both controlled environments and real-world scenarios.

\label{section: References}
\bibliographystyle{IEEEtran}
\bibliography{references.bib}

\begin{thebibliography}{10}
\providecommand{\url}[1]{#1}
\csname url@samestyle\endcsname
\providecommand{\newblock}{\relax}
\providecommand{\bibinfo}[2]{#2}
\providecommand{\BIBentrySTDinterwordspacing}{\spaceskip=0pt\relax}
\providecommand{\BIBentryALTinterwordstretchfactor}{4}
\providecommand{\BIBentryALTinterwordspacing}{\spaceskip=\fontdimen2\font plus
\BIBentryALTinterwordstretchfactor\fontdimen3\font minus \fontdimen4\font\relax}
\providecommand{\BIBforeignlanguage}[2]{{%
\expandafter\ifx\csname l@#1\endcsname\relax
\typeout{** WARNING: IEEEtran.bst: No hyphenation pattern has been}%
\typeout{** loaded for the language `#1'. Using the pattern for}%
\typeout{** the default language instead.}%
\else
\language=\csname l@#1\endcsname
\fi
#2}}
\providecommand{\BIBdecl}{\relax}
\BIBdecl

\bibitem{tiktok}
\BIBentryALTinterwordspacing
``Tiktok: Bipedal robot.'' [Online]. Available: \url{http://ruina.tam.cornell.edu/research/topics/locomotion_and_robotics/Tik-Tok/index.html}
\BIBentrySTDinterwordspacing

\bibitem{bolt}
\BIBentryALTinterwordspacing
Open-Dynamic-Robot-Initiative, ``Open robot actuator hardware.'' [Online]. Available: \url{https://github.com/open-dynamic-robot-initiative/open_robot_actuator_hardware}
\BIBentrySTDinterwordspacing

\bibitem{cassie}
J.~Reher, W.-L. Ma, and A.~D. Ames, ``Dynamic walking with compliance on a cassie bipedal robot,'' in \emph{2019 18th European Control Conference (ECC)}, 2019, pp. 2589--2595.

\bibitem{Apptronik}
\BIBentryALTinterwordspacing
 [Online]. Available: \url{https://apptronik.com/our-work}
\BIBentrySTDinterwordspacing

\bibitem{park2001impedance}
J.~H. Park, ``Impedance control for biped robot locomotion,'' \emph{IEEE Transactions on Robotics and Automation}, vol.~17, no.~6, pp. 870--882, 2001.

\bibitem{lim2004position}
H.-O. Lim, S.~A. Setiawan, and A.~Takanishi, ``Position-based impedance control of a biped humanoid robot,'' \emph{Advanced Robotics}, vol.~18, no.~4, pp. 415--435, 2004.

\bibitem{santacruz2013reactive}
C.~Santacruz and Y.~Nakamura, ``Reactive stepping strategies for bipedal walking based on neutral point and boundary condition optimization,'' in \emph{2013 IEEE International Conference on Robotics and Automation}.\hskip 1em plus 0.5em minus 0.4em\relax IEEE, 2013, pp. 3110--3115.

\bibitem{raibert1985legged}
M.~H. Raibert and E.~R. Tello, ``Legged robots that balance,'' \emph{IEEE Expert}, vol.~1, no.~4, pp. 89--89, 1986.

\bibitem{apgar2018fast}
T.~Apgar, P.~Clary, K.~Green, A.~Fern, and J.~W. Hurst, ``Fast online trajectory optimization for the bipedal robot cassie.'' in \emph{Robotics: Science and Systems}, vol. 101.\hskip 1em plus 0.5em minus 0.4em\relax Pittsburgh, Pennsylvania, USA, 2018, p.~14.

\bibitem{ferreau2008online}
H.~J. Ferreau, H.~G. Bock, and M.~Diehl, ``An online active set strategy to overcome the limitations of explicit mpc,'' \emph{International Journal of Robust and Nonlinear Control: IFAC-Affiliated Journal}, vol.~18, no.~8, pp. 816--830, 2008.

\bibitem{kajita20013d}
S.~Kajita, F.~Kanehiro, K.~Kaneko, K.~Yokoi, and H.~Hirukawa, ``The 3d linear inverted pendulum mode: A simple modeling for a biped walking pattern generation,'' in \emph{Proceedings 2001 IEEE/RSJ International Conference on Intelligent Robots and Systems. Expanding the Societal Role of Robotics in the the Next Millennium (Cat. No. 01CH37180)}, vol.~1.\hskip 1em plus 0.5em minus 0.4em\relax IEEE, 2001, pp. 239--246.

\bibitem{westervelt2003hybrid}
E.~R. Westervelt, J.~W. Grizzle, and D.~E. Koditschek, ``Hybrid zero dynamics of planar biped walkers,'' \emph{IEEE transactions on automatic control}, vol.~48, no.~1, pp. 42--56, 2003.

\bibitem{ding2022orientation}
Y.~Ding, C.~Khazoom, M.~Chignoli, and S.~Kim, ``Orientation-aware model predictive control with footstep adaptation for dynamic humanoid walking,'' in \emph{2022 IEEE-RAS 21st International Conference on Humanoid Robots (Humanoids)}.\hskip 1em plus 0.5em minus 0.4em\relax IEEE, 2022, pp. 299--305.

\bibitem{tassa2012synthesis}
Y.~Tassa, T.~Erez, and E.~Todorov, ``Synthesis and stabilization of complex behaviors through online trajectory optimization,'' in \emph{2012 IEEE/RSJ International Conference on Intelligent Robots and Systems}.\hskip 1em plus 0.5em minus 0.4em\relax IEEE, 2012, pp. 4906--4913.

\bibitem{kuindersma2016optimization}
S.~Kuindersma, R.~Deits, M.~Fallon, A.~Valenzuela, H.~Dai, F.~Permenter, T.~Koolen, P.~Marion, and R.~Tedrake, ``Optimization-based locomotion planning, estimation, and control design for the atlas humanoid robot,'' \emph{Autonomous robots}, vol.~40, pp. 429--455, 2016.

\bibitem{kuindersma2020recent}
S.~Kuindersma, ``Recent progress on atlas, the world’s most dynamic humanoid robot,'' \emph{Robotics Today-A Series of Technical Talks.[Online]. Available: https://youtu. be/EGABAx52GKI}, 2020.

\bibitem{sutton2018reinforcement}
R.~S. Sutton and A.~G. Barto, \emph{Reinforcement learning: An introduction}.\hskip 1em plus 0.5em minus 0.4em\relax MIT press, 2018.

\bibitem{heess2017emergence}
N.~Heess, D.~Tb, S.~Sriram, J.~Lemmon, J.~Merel, G.~Wayne, Y.~Tassa, T.~Erez, Z.~Wang, S.~Eslami \emph{et~al.}, ``Emergence of locomotion behaviours in rich environments,'' \emph{arXiv preprint arXiv:1707.02286}, 2017.

\bibitem{benbrahim1997biped}
H.~Benbrahim and J.~A. Franklin, ``Biped dynamic walking using reinforcement learning,'' \emph{Robotics and Autonomous systems}, vol.~22, no. 3-4, pp. 283--302, 1997.

\bibitem{xie2018feedback}
Z.~Xie, G.~Berseth, P.~Clary, J.~Hurst, and M.~van~de Panne, ``Feedback control for cassie with deep reinforcement learning,'' in \emph{2018 IEEE/RSJ International Conference on Intelligent Robots and Systems (IROS)}.\hskip 1em plus 0.5em minus 0.4em\relax IEEE, 2018, pp. 1241--1246.

\bibitem{siekmann2021sim}
J.~Siekmann, Y.~Godse, A.~Fern, and J.~Hurst, ``Sim-to-real learning of all common bipedal gaits via periodic reward composition,'' in \emph{2021 IEEE International Conference on Robotics and Automation (ICRA)}.\hskip 1em plus 0.5em minus 0.4em\relax IEEE, 2021, pp. 7309--7315.

\bibitem{siekmann2021blind}
J.~Siekmann, K.~Green, J.~Warila, A.~Fern, and J.~Hurst, ``Blind bipedal stair traversal via sim-to-real reinforcement learning,'' \emph{arXiv preprint arXiv:2105.08328}, 2021.

\bibitem{li2021reinforcement}
Z.~Li, X.~Cheng, X.~B. Peng, P.~Abbeel, S.~Levine, G.~Berseth, and K.~Sreenath, ``Reinforcement learning for robust parameterized locomotion control of bipedal robots,'' in \emph{2021 IEEE International Conference on Robotics and Automation (ICRA)}.\hskip 1em plus 0.5em minus 0.4em\relax IEEE, 2021, pp. 2811--2817.

\bibitem{kumar2022adapting}
A.~Kumar, Z.~Li, J.~Zeng, D.~Pathak, K.~Sreenath, and J.~Malik, ``Adapting rapid motor adaptation for bipedal robots,'' in \emph{2022 IEEE/RSJ International Conference on Intelligent Robots and Systems (IROS)}.\hskip 1em plus 0.5em minus 0.4em\relax IEEE, 2022, pp. 1161--1168.

\bibitem{castillo2022reinforcement}
G.~A. Castillo, B.~Weng, W.~Zhang, and A.~Hereid, ``Reinforcement learning-based cascade motion policy design for robust 3d bipedal locomotion,'' \emph{IEEE Access}, vol.~10, pp. 20\,135--20\,148, 2022.

\bibitem{9682564}
L.~Krishna, G.~A. Castillo, U.~A. Mishra, A.~Hereid, and S.~Kolathaya, ``Linear policies are sufficient to realize robust bipedal walking on challenging terrains,'' \emph{IEEE Robotics and Automation Letters}, vol.~7, no.~2, pp. 2047--2054, 2022.

\bibitem{paigwar2021robust}
K.~Paigwar, L.~Krishna, S.~Bhatnagar, A.~Ghosal, B.~Amrutur, S.~Kolathaya \emph{et~al.}, ``Robust quadrupedal locomotion on sloped terrains: A linear policy approach,'' in \emph{Conference on Robot Learning}.\hskip 1em plus 0.5em minus 0.4em\relax PMLR, 2021, pp. 2257--2267.

\bibitem{rahme2021linear}
M.~Rahme, I.~Abraham, M.~L. Elwin, and T.~D. Murphey, ``Linear policies are sufficient to enable low-cost quadrupedal robots to traverse rough terrain,'' in \emph{2021 IEEE/RSJ International Conference on Intelligent Robots and Systems (IROS)}.\hskip 1em plus 0.5em minus 0.4em\relax IEEE, 2021, pp. 8469--8476.

\bibitem{krishna2021learning}
L.~Krishna, U.~A. Mishra, G.~A. Castillo, A.~Hereid, and S.~Kolathaya, ``Learning linear policies for robust bipedal locomotion on terrains with varying slopes,'' in \emph{2021 IEEE/RSJ International Conference on Intelligent Robots and Systems (IROS)}.\hskip 1em plus 0.5em minus 0.4em\relax IEEE, 2021, pp. 5159--5164.

\bibitem{opendynamic}
\BIBentryALTinterwordspacing
F.~Grimminger, A.~Meduri, M.~Khadiv, J.~Viereck, M.~W{\"u}thrich, M.~Naveau, V.~Berenz, S.~Heim, F.~Widmaier, T.~Flayols, J.~Fiene, A.~Badri-Spr{\"o}witz, and L.~Righetti, ``An open torque-controlled modular robot architecture for legged locomotion research,'' \emph{IEEE Robotics and Automation Letters}, vol.~5, no.~2, pp. 3650--3657, Apr. 2020. [Online]. Available: \url{https://arxiv.org/abs/1910.00093}
\BIBentrySTDinterwordspacing

\bibitem{ekf}
S.~Bonnable, P.~Martin, and E.~Salaün, ``Invariant extended kalman filter: theory and application to a velocity-aided attitude estimation problem,'' in \emph{Proceedings of the 48h IEEE Conference on Decision and Control (CDC) held jointly with 2009 28th Chinese Control Conference}, 2009, pp. 1297--1304.

\bibitem{DBLP:journals/corr/abs-1803-07055}
\BIBentryALTinterwordspacing
H.~Mania, A.~Guy, and B.~Recht, ``Simple random search provides a competitive approach to reinforcement learning,'' \emph{CoRR}, vol. abs/1803.07055, 2018. [Online]. Available: \url{http://arxiv.org/abs/1803.07055}
\BIBentrySTDinterwordspacing

\bibitem{todorov2012mujoco}
E.~Todorov, T.~Erez, and Y.~Tassa, ``Mujoco: A physics engine for model-based control,'' in \emph{2012 IEEE/RSJ international conference on intelligent robots and systems}.\hskip 1em plus 0.5em minus 0.4em\relax IEEE, 2012, pp. 5026--5033.

\end{thebibliography}

\end{document}